\documentclass[conference]{IEEEtran}
\IEEEoverridecommandlockouts
\usepackage{cite}
\usepackage{amsmath,amssymb,amsfonts}
\usepackage{algorithmic}
\usepackage{graphicx}
\graphicspath{{Figures/}}
\usepackage{textcomp}
\usepackage{xcolor}
\usepackage{comment}
\usepackage[utf8]{inputenc}
\def\BibTeX{{\rm B\kern-.05em{\sc i\kern-.025em b}\kern-.08em
    T\kern-.1667em\lower.7ex\hbox{E}\kern-.125emX}}

\begin{document}

\title{A Comparative Analysis of Machine Learning Models for Long and Short-Term Forecasting of the Egyptian Stock Market: A Focus on EGX30\\}
\author{\IEEEauthorblockN{Anonymous}}

\author{\IEEEauthorblockN{Muhammed Walid$^*$, Ahmed El-Naeimy$^*$, Hosam Moubarak$^{\dag}$, Walid Gomaa$^{*,+}$}

\IEEEauthorblockA{$^*$ Cyber Physical Lab, Department of Computer Science \& Engineering\\
\dag Accounting Department\\
Egypt-Japan University Of Science and Technology, Alexandria, Egypt}
\IEEEauthorblockA{$^+$ Faculty of Engineering, Alexandria University, 
Alexandria, Egypt}
\IEEEauthorblockA{\{mohamed.waleed, ahmed.eelnaeimy, hosam.moubarak, walid.gomaa\}@ejust.edu.eg}}
\maketitle

\begin{abstract} \label{abstract}
This study concentrates on predicting stock prices in the Egyptian market, focusing on the EGX30, an influential financial hub in the Middle East. While most research focuses on global stocks, there's a growing need to understand stock trends in developing countries like Egypt. The study compares different machine learning models for forecasting EGX30 trends, covering short and long-term predictions. Using historical EGX30 data, including metrics like root mean squared error, Mean Absolute Percentage Error, and coefficient of determination, models like K-Nearest Neighbours, random forest, extreme gradient boosting, long short-term memory networks, and gated recurrent unit networks were evaluated. The goal is to determine the most effective models for EGX30 prediction, considering Egypt's unique market dynamics. Insights from this study aid investors in making informed decisions. Results show that the Gated Recurrent Unit (GRU) outperformed the other models in the one-week, one-month, and two-months while the eXtreme Gradient Boosting (XGBoost) model outperformed others in the one-day predictions, highlighting their usefulness in predictive analysis for financial markets. The study also showed the importance of using the ensemble techniques, especially in the long-term predictions which proved better results reaching 5 times the GRU in the two-month predictions. Additionally, the study notes the surprisingly good performance of K-Nearest Neighbours (KNN) on long-term predictions, suggesting its enduring relevance and potential for future applications in the fintech domains.

\end{abstract}

\begin{IEEEkeywords}
Stock Market Prediction, Financial Time Series, Machine Learning, Deep Learning, Technical Analysis, Egyptian Stock Market

\end{IEEEkeywords}

\section{Introduction} \label{Introduction}
The Egyptian stock market, an emerging financial arena, is marked by ongoing expansion and presents various prospects for experienced investors and newcomers alike~\cite{elmosalamy2018predictors}. As the market progresses, it offers a dynamic platform for trading and investment. However, the potential for high returns is accompanied by inherent risks that may pose challenges for investors with limited trading knowledge. 
Moreover, a significant obstacle lies in the lack of technical analysis expertise among regular traders and investors in this market, highlighting a broader need for financial literacy \cite{zhou2023market}. In light of these circumstances, this investigation seeks to bridge this gap by harnessing the predictive capabilities of advanced machine learning and deep learning models to provide precise predictions tailored specifically for the Egyptian stock market. The models are built with a strong structure and flexibility, making it easy to adapt them to different stock markets with minimal changes. Despite being trained on specific datasets, the model's core principles and adaptable design allow it to effectively work with new datasets that share similar characteristics.
\par 
This research aims to empower investors, reduce risks, and contribute to overall economic growth within this expanding financial domain.
Accurately predicting trends in the financial markets has become increasingly crucial in recent years due to its applications ranging from risk management to investment strategy formulation~\cite{Luo2022}. 
The complex nature of financial markets calls for advanced predictive models capable of capturing intricate patterns within time-series data~\cite{wen2019stock}. This study focuses on utilizing machine learning and deep learning techniques with the objective of enhancing predictive capabilities pertinent specifically to forecasting both short-term as well as long-term trends within the Egyptian stock market.
Six distinct machine learning models in this research have been chosen: Extreme Gradient Boosting, Random Forests, K-Nearest Neighbour, Extra Trees, and Adaptive Boosting along Light Gradient Boosted Machine. 
Two versions of predictive models were employed; one was designed for short term predictions  while another was configured for foreseeing long term predictions where in short-term predictions extended up until 1 week into the future while long-term projections spanned over periods up to two months. 
The accuracy of predictions was enhanced by considering past values in this study. Improvement was achieved through the utilization of a lag function on historical data, enabling the expectation of trends over multiple days. The inherent temporal structure present in various financial data types was thereby respected through this approach, contributing to an elevated level of forecast reliability. \cite{Laws2018}. 

\par 
In addition to traditional machine learning (ML) models, the analysis included leveraging deep-learning techniques via Long Short-Term Memory and Gated Recurrent Unit. This incorporation of the advanced neural network models designed to capture complex patterns found in financial time-series data.~\cite{Hansika_2021}. 
A unique contribution made by this study is incorporating an ensemble of top-performing models to ensure more accurate results that can accommodate evolving market conditions. This ensemble model applies the blending technique with the best-performing models which gives this model more power and enhances its robustness~\cite{Iliopoulos_2023}. The comprehensive goal of this model is to achieve the utmost precision in forecasting Egyptian stock market trends both in the short term and long term. 
The rest of the paper is organized as follows:
\ref{Literature Review} Literature Review that surveys existing research on machine learning in financial markets, tracing the evolution of predictive models. \ref{Methodology} The Methodology outlines the research design, data collection, and preprocessing steps, introducing selected machine learning and deep learning models along with technical indicators and preprocessing techniques. \ref{Experimental Results} In Experimental Results, the outcomes of various models, including traditional machine learning and deep learning architectures, are presented and analyzed. \ref{Conclusion} The Conclusion and \ref{Limitations and Futre Work} Limitations and Futre Work summarize key findings, discusses implications, suggests future research directions, and reflects on study limitations while offering recommendations for practitioners and researchers in financial market prediction.

\section{Literature Review} \label{Literature Review}
In recent years, the intersection of machine learning and finance has witnessed a surge in research, particularly in predicting stock market trends~\cite{Addagalla_2023}. As scholars continue to delve into the intricacies of predictive analytics~\cite{Massaad_2019}, ongoing research focuses on improving the accuracy, interpretability, and fairness of predictive models~\cite{hardt_2016}. This can be achieved by exploring diverse prediction models, dissecting their application domains, and scrutinizing innovative data sources~\cite{Halevy_2009}. In doing so, it contributes to a nuanced understanding of the evolving landscape in financial forecasting~\cite{Makridakis_2018}.
 
\subsection{Predictive Models and Techniques in Financial Markets}

\par 
In the realm of financial markets, predictive models serve as indispensable tools, employing sophisticated statistical and mathematical techniques to proactively anticipate forthcoming market dynamics and asset valuations~\cite{Daniel_2022}. These models are made to carefully analyze past data, find detailed patterns, and help people make informed decisions.
The study in~\cite{Daniel_2022} undertakes a comprehensive comparative analysis that encompasses various machine learning (ML) and deep learning (DL) models, with a particular emphasis on the pivotal role of input data representation and the nuanced evaluation of performance metrics.

\par 
Significant findings underscore the ascendancy of deep learning methods, specifically Recurrent Neural Networks (RNN) and Long Short-Term Memory (LSTM), surpassing conventional ML models in terms of efficacy. This observation is corroborated in~\cite{Mukherjee_2021}, which proposes models utilizing solely deep learning models for stock market predictive purposes.
However,~\cite{Khanna_2022} diverges by employing four distinct mixture set of classical and deep models—namely, Long Short-Term Memory, XGBoost, Support Vector Machine, and Random Forests. Intriguingly, the highest accuracy is attributed to Random Forests, deviating from the findings in both~\cite{Nabipour_2020} and~\cite{Mukherjee_2021}.
In the execution of the previous analytical perspective, a diverse array of models is applied to the proposed study , concluded in the formulation of an ensemble model. This ensemble model merging the optimal three models for each prediction period, synthesizing both deep learning and machine learning models, all while incorporating technical analysis as a pivotal component.
This multifaceted approach enhances the robustness of the predictive framework, thereby contributing to a more nuanced and comprehensive understanding of market dynamics.

\subsection{Data Preprocessing Impact}

\par 
In the examination of the influence of data preprocessing techniques on the prediction of stock market trends, the authors in~\cite{Nabipour_2020} employ two distinct methodologies. The first approach involves continuous data 
manipulation, wherein technical indicators such as Simple Moving Average (SMA) and Weighted Moving Average (WMA) are computed based on stock trading values. The second approach, referred to as the binary data method, entails the conversion of these indicators which reveales underlying patterns crucial for informed decision-making and reliable predictive modeling.
In the study~\cite{Khanna_2022}, ten technical indicators 
were proposed, comprising nine identical indicators to those in~\cite{Nabipour_2020} (SMA, WMA, Momentum (MOM), Stochastic Oscillator (STC), Stochastic \%K (STK), Relative Strength Index (RSI), Larry Williams \%R (LWR), Accumulation Distribution Oscillator (ADO), and Commodity Channel Index (CCI)). 
The only divergence lies in the utilization of Signal Line (SIG) in~\cite{Nabipour_2020}, while~\cite{Khanna_2022} employs Moving Average Convergence Divergence (MACD). 

\par 
In alignment with this, This work incorporates these eleven indicators in the preprocessing phase,
introducing additional elements such as Bollinger Bands, Exponential Moving Average (EMA) instead of the WMA. To enhance the predictive capabilities of our models, we introduce the use of lag features and rolling windows. By incorporating historical stock market data through these mechanisms, it provided our models with a broader temporal context, allowing for a more nuanced understanding of trends and patterns.

\subsection{Prediction Intervals}

\par 
In the study~\cite{Mukherjee_2021}, the primary emphasis was placed on optimizing short-term predictions to achieve maximal accuracy with minimal error. The utilization of predictive models involved the strategic forecasting of forthcoming data values based on preceding days data points. 
This recursive forecasting process was iteratively applied, contingent upon the continued validity of the dataset under consideration.
It is noteworthy, as expounded in~\cite{Khanna_2022}, that superior outcomes, measured in terms of both accuracy and F1-score, were attained through the implementation of classification models for long-term predictions within 
the temporal range of 17-20 days. This finding stands in contrast to the results in~\cite{Mukherjee_2021}, where optimal performance was observed in the context of single-day predictions. Consequently, predictions for a singular day ahead, short-term horizons spanning from one day to one week , and long-term intervals spanning from one month to two months were encompassed in this proposed investigation.
This comprehensive approach allowed for a comparative analysis aimed at discerning not only the highest achievable accuracy, but also the identification of the most precise trends—a pivotal consideration within the domain of stock market analysis~\cite{Shen_2020}.
This dual-focus approach, grounded in both short-term and long-term prediction assessments, contributes to a more nuanced understanding of the intricacies inherent in forecasting financial markets.

\subsection{Summary}

\par 
This review provides a comprehensive analysis of the intersection between machine learning and finance, focusing particularly on predicting stock market trends. It examines the prevalence of deep learning compared to traditional models across various datasets, highlighting instances where deep learning outperforms traditional methods and vice versa.
An in-depth analysis of data preprocessing techniques is conducted, including the use of technical indicators and supplementary elements like Bollinger Bands to enhance predictive capabilities. The importance of tailoring models to specific time horizons is emphasized through the exploration of prediction intervals, covering both short-term predictions (1 day- 1 week) and long-term forecasts (1-2 months).


\section{Methodology} \label{Methodology}

\par 
A diverse array of seven machine learning algorithms was selected for analysis. These included K-Nearest Neighbours (KNN), Decision Tree Regressor, Random Forest, Extra Trees, Extreme Gradient Boosting (XGBoost), Adaptive Boosting, and Light Gradient Boosting Machine (LightGBM). In the realm of deep learning, Long Short-Term Memory (LSTM) and Gated Recurrent Unit (GRU) models were also incorporated. This selection was aimed at capturing the complex patterns and dependencies present in financial time series data \cite{Touzani_2021}.
Extensive feature engineering was conducted to enhance the predictive capabilities of the models. This process involved the systematic analysis and selection of financial indicators critical for capturing the essence of market dynamics. In assessing the performance of these models, a rolling forecast methodology was employed. Key evaluation metrics, such as the root mean squared error (RMSE) and other relevant statistical measures, were utilized to ensure an exhaustive evaluation of each model's predictive accuracy.
comprehensibility analyses were conducted for each model. This was crucial in providing transparent and actionable insights to stakeholders such as investors, traders, and financial analysts.

\subsection{The Data }
\subsubsection{EGX30}
The EGX30 index, the leading benchmark for the Egyptian Stock Exchange, comprises the top 30 actively traded stocks on the EGX and is acknowledged as a key indicator of the Egyptian stock market's overall performance \cite{Neveen_2018}. It provides substantial insights into market trends and investor sentiment within Egypt. The index's inclusion in this analysis is justified by its extensive monitoring by institutional investors, fund managers, and other market participants, rendering it an essential reference for evaluating investment portfolio performance, assessing market conditions, and informing investment decisions \cite{yang_2017}.
Incorporating this indexes into the study enhances the understanding of market dynamics, allowing for the development of informed investment strategies and effective risk management. The EGX30's broad sector representation aids in identifying specific opportunities and comprehending market movements, crucial for portfolio diversification \cite{ibrahim_2019}.The index's reflection of market sentiment and trends assists in discerning investor confidence and anticipating market shifts, while its response to economic indicators and policy decisions offers insights into the complex interplay between market dynamics and broader economic factors \cite{sun_2016}.
\subsubsection{Preprocessing}
\begin{figure}
    \centering
    \includegraphics[width=0.75\linewidth]{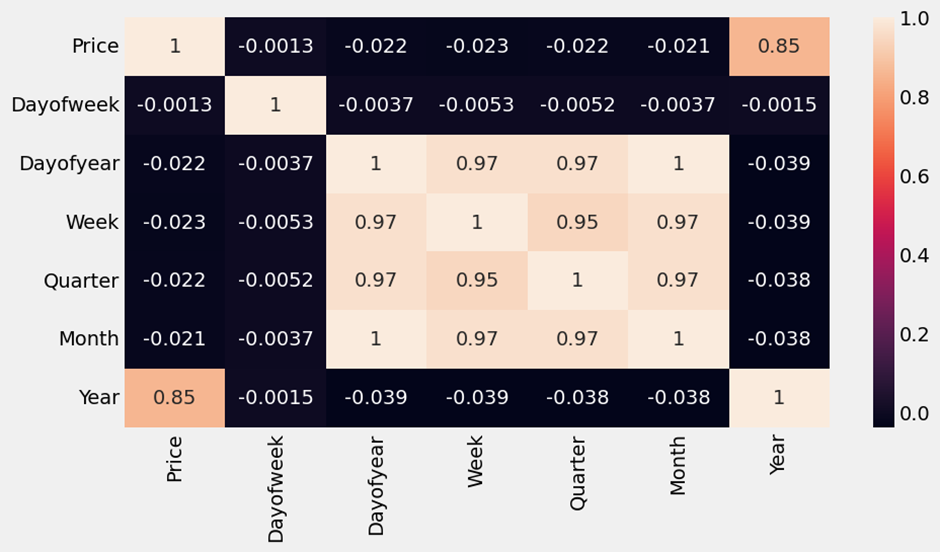}
    \caption{Correlation Matrix Highlighting the Relationships Between Stock Prices and (Dayofweek, Dayofyear, Week, Quarter, Month, Year}
    \label{co_mat_simple}
\end{figure}

In the preliminary phase of the analysis, the dataset was segmented into various temporal features: day of week, day of year, Week, Quarter, Month, and Year. Evaluation of the correlation matrix a statistical representation of pairwise correlation coefficients between variables, exhibited a low correlation between stock prices and the selected temporal features. Additionally, high correlations among some features were observed. Fig.~\ref{co_mat_simple} revealed that these features alone were insufficient for accurate stock price prediction due to the multicollinearity issues.
To mitigate these limitations, rolling window lag features were introduced, enabling the model to incorporate historical data from the preceding months. This approach was aimed at providing the model with a deeper historical context. Furthermore, the rolling window lag technique served as a temporal aggregation method essential for capturing historical trends, seasonal variations, and cyclic patterns in the data \cite{Atsushi_2017}. The incorporation of this technique allows the model to assimilate current trends and predict future patterns with enhanced accuracy.
\subsubsection{Technical Analysis}
\begin{figure}[!h]
    \centering
    \includegraphics[width=0.75\linewidth]{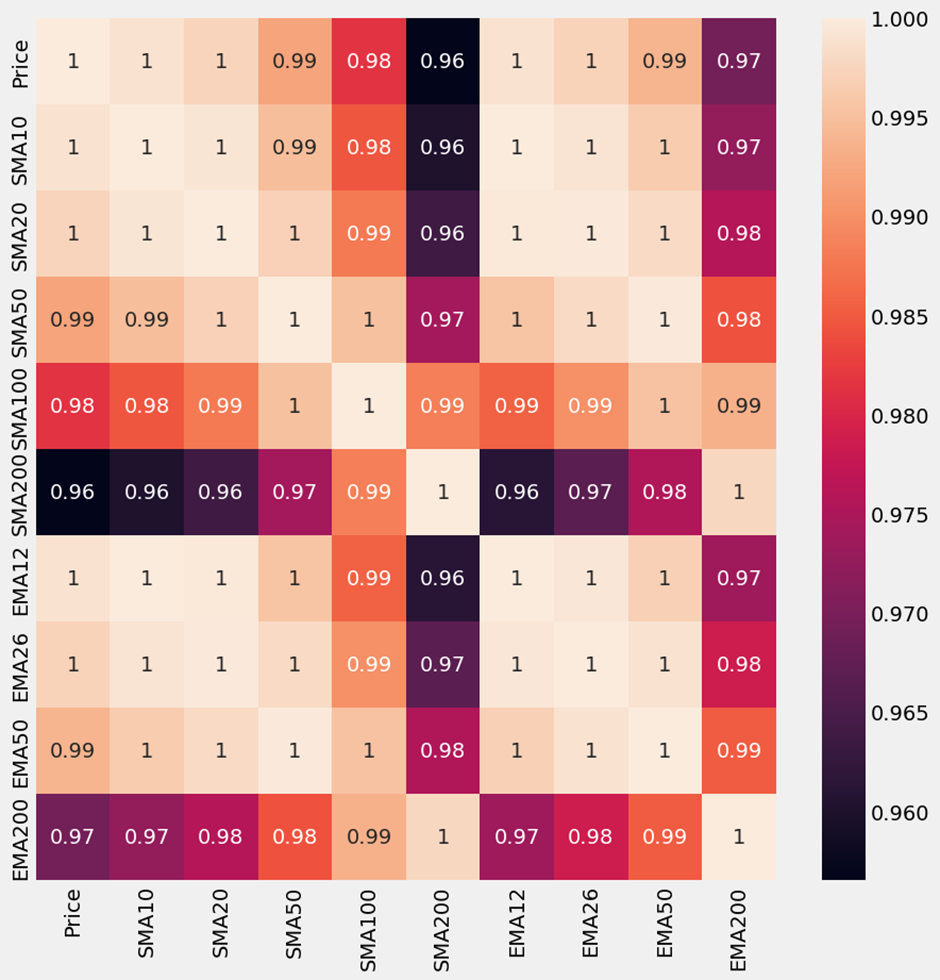}
    \caption{Detailed Analysis of Correlation Between Rolling Window Features and Stock Prices, Underlining Their Importance in Predictive Accuracy}
    \label{co_mat_sma_ema}
\end{figure}

In the feature engineering process of this analysis, we used 9 technical indexes two key metrics were the Simple Moving Average (SMA) and the Exponential Moving Average (EMA). The SMA is computed as the average of a selected range of values, providing a basic representation of trends over a specified period. In contrast, the EMA assigns greater weight to recent data, making it more sensitive to short-term market movements. \cite{desouza_2018}
With the inclusion of lag features and a rolling window, it was observed that the lag features exhibited substantial correlations with one another, suggesting the need for a reduction in the number of lags to optimize the model. Additionally, the features derived from the rolling window techniques (SMAs and EMAs) showed significant correlations with the target variable, indicating their crucial role in the predictive models.
Further analysis of the correlation between SMAs, EMAs, and stock prices Fig.~\ref{co_mat_sma_ema} confirmed the essential contribution of these features in enhancing the accuracy of stock price predictions.
Other indicators like CCI and RSI were used as other studies showed in the literature review these indicators proved their ability to predict market trends.
\subsection{Traditional Machine Learning Models}
The Traditional Machine Learning Models form a bedrock rooted in established statistical principles and algorithmic frameworks. This foundation enables the understanding of complex patterns and connections within historical market data \cite{wang_2021}.
\subsubsection{K-Nearest Neighbors (KNN)}
K-Nearest Neighbors (KNN) is a non-parametric and instance-based algorithm for classification and regression.\cite{Samya_2021} For prediction, it identifies the k-nearest data points in the feature space and assigns the majority class (for classification) or the average value (for regression).
\subsubsection{Decision Tree Regressor}
Decision Tree Regressor recursively splits the dataset based on features to create a tree-like structure. Each split is determined by maximizing the reduction in variance or mean squared error, resulting in a series of decisions leading to a predicted value \cite{Houssainy_2021}.
\subsubsection{Random Forest}
Random Forest is an ensemble method that constructs multiple decision trees. It aggregates predictions by averaging (for regression) or voting (for classification) across the individual trees, thereby reducing overfitting \cite{Random_Forests}.
\subsubsection{Extra Trees}
Extra Trees, like Random Forest, construct multiple decision trees. However, it selects random thresholds for each feature at every split, introducing additional randomness \cite{Extra_Trees}.

\subsection{Gradient Boosting Models}
Gradient Boosting Models represent a sophisticated evolution in predictive analytics. Founded on the principles of boosting, these models iteratively refine predictions, leveraging the strengths of decision trees. By emphasizing model weaknesses in each iteration, Gradient Boosting cultivates an adaptive learning process that excels in capturing nuanced relationships within historical market data \cite{Gradient_Boosting}.
\subsubsection{Extreme Gradient Boosting (XGBoost)}
XGBoost is a gradient-boosting algorithm that minimizes a cost function by sequentially adding weak learners. It combines the predictions of multiple trees, placing more emphasis on instances with higher prediction errors \cite{XGBoost}. 
\subsubsection{Adaptive Boosting}
Adaptive Boosting (AdaBoost) combines multiple weak learners into a strong model. It assigns higher weights to misclassified instances, enabling subsequent weak learners to focus on correcting errors \cite{Adaptive_Boosting}.
\subsubsection{Light Gradient Boosting Machine (LightGBM)}
LightGBM is a gradient-boosting framework designed for efficiency. It builds trees in a depth-wise manner, prioritizing leaf nodes with higher gradients for better optimization \cite{Athiyarath_2020}. 
\subsection{Deep Learning Models}
The Deep Learning Models models serve as advanced analytical agents, meticulously crafted to unravel intricate patterns. With a primary emphasis on memory retention and adaptive learning, these models demonstrate exceptional proficiency in discerning complex, time-sensitive market dynamics. Moreover, they exhibit an unparalleled capacity to capture and comprehend temporal dependencies, thereby contributing significantly to the nuanced analysis of intricate phenomena within various domains \cite{Ilya_2014}. 
\subsubsection{Long Short-Term Memory (LSTM)}
Long Short-Term Memory (LSTM) is a specialized variant of recurrent neural networks (RNNs) designed to address long-term dependencies in sequential data. LSTMs introduce memory cells for selective information storage, mitigating the vanishing gradient problem and enhancing their ability to comprehend intricate patterns. This makes LSTMs well-suited for tasks involving prolonged temporal dependencies \cite{LSTM}.

\subsubsection{Gated Recurrent Unit (GRU)}
The Gated Recurrent Unit (GRU) is a type of recurrent neural network, akin to LSTM, designed for efficient learning of sequential patterns. It utilizes gating mechanisms to regulate information flow, striking a balance between simplicity and performance compared to LSTM. GRU is proficient in capturing and utilizing temporal dependencies in sequential data \cite{GRU}.

\subsection{Optimizers and Accuracy Metrics}
Optimizers and accuracy metrics are integral elements in the refinement and evaluation of predictive models. In the domain of deep learning, the Adam optimizer assumes a central role by employing adaptive learning rates to enhance convergence. Concurrently, the Root Mean Squared Error (RMSE) stands out as a crucial metric for assessing model performance, proficiently quantifying average prediction errors. Together, these components play a pivotal role in guiding the learning process, influencing convergence, and determining a model's overall effectiveness and reliability \cite{Zhang_2016}.
\subsubsection{Adam Optimizer}
Adam combines the benefits of two other popular optimizers, namely AdaGrad and RMSProp, by maintaining adaptive learning rates for each parameter. This adaptability proves crucial in scenarios where different model parameters require different update magnitudes for effective convergence. The algorithm's ability to automatically adjust learning rates accelerates convergence and mitigates the challenges associated with manual tuning. The adaptive nature of Adam ensures that the model navigates through complex loss landscapes efficiently, making it a preferred choice for optimizing deep learning models. The efficient convergence facilitated by the Adam optimizer contributes to faster training times and an increased likelihood of discovering optimal model parameters \cite{Adam}.
\subsubsection{Root Mean Squared Error (RMSE)}
The RMSE provides a comprehensive measure of the average prediction errors by considering both the magnitude and direction of deviations between predicted and actual values. The square root operation in RMSE ensures that larger errors are appropriately penalized, making it particularly sensitive to outliers. This sensitivity is valuable in scenarios where accurate quantification of prediction accuracy is essential. RMSE is easily interpretable, as it represents the standard deviation of the residuals, providing a clear understanding of the model's overall performance \cite{Wang_2018}. \[RMSE= \sqrt{\frac{1}{n}\sum_{i=1}^{n} (\hat{y_i}-y_i)^2}\]
\subsubsection{Mean Absolute Percentage Error (MAPE)}
The Mean Absolute Percentage Error (MAPE) is a widely used metric for evaluating the accuracy of predictive models. It quantifies the average percentage difference between predicted and actual values. MAPE is particularly valuable for understanding the scale of errors in predictions, as it directly measures the relative size of the errors in percentage terms. MAPE provides a clear and interpretable measure of the average percentage deviation between predicted and actual values. It is especially useful when a straightforward understanding of prediction accuracy, in terms of percentage error, is crucial for decision-making \cite{KIM_2016}. \[MAPE= \frac{1}{n}\sum_{i=1}^{n}|\frac{\hat{y_i}-y_i}{y_i}| \times 100\]
\subsubsection{coefficient of determination (R2)}
The R2 score, also known as the coefficient of determination, is a statistical metric used to assess how well a regression model explains the variability in the dependent variable based on the independent variables. Ranging from 0 to 1, where 0 signifies no explanatory power and 1 signifies perfect prediction, the R2 score provides insights into the model's ability to capture data variability. A higher R2 score, closer to 1, indicates a better fit, suggesting that the model effectively explains the data. Conversely, a lower R2 score, closer to 0, implies a poor fit, indicating the model inadequately accounts for the data's variability \cite{Wang_2017}.
\[R2 = 1-\frac{\sum_{i=1}^{n}(y_i-\hat{y_i})^2}{\sum_{i=1}^{n}(y_i-\Bar{y})^2}\]
\section{Experimental Results} \label{Experimental Results}
The predictive modeling performance of diverse machine learning algorithms in forecasting the closing prices of the EGX30 index on the Egyptian Stock Exchange was rigorously examined in this study. The aim was to contribute insights to the domain of predictive modeling in finance. Evaluations of model performance were conducted for both short-term (one day and one week) and long-term (one month and two months) prediction periods. The results are presented to enhance the understanding of predictive modeling in financial contexts.
\subsection{Short Term Predictions}
\subsubsection{One-Day Predictions}
\begin{figure}
    \centering
    \includegraphics[width=1\linewidth]{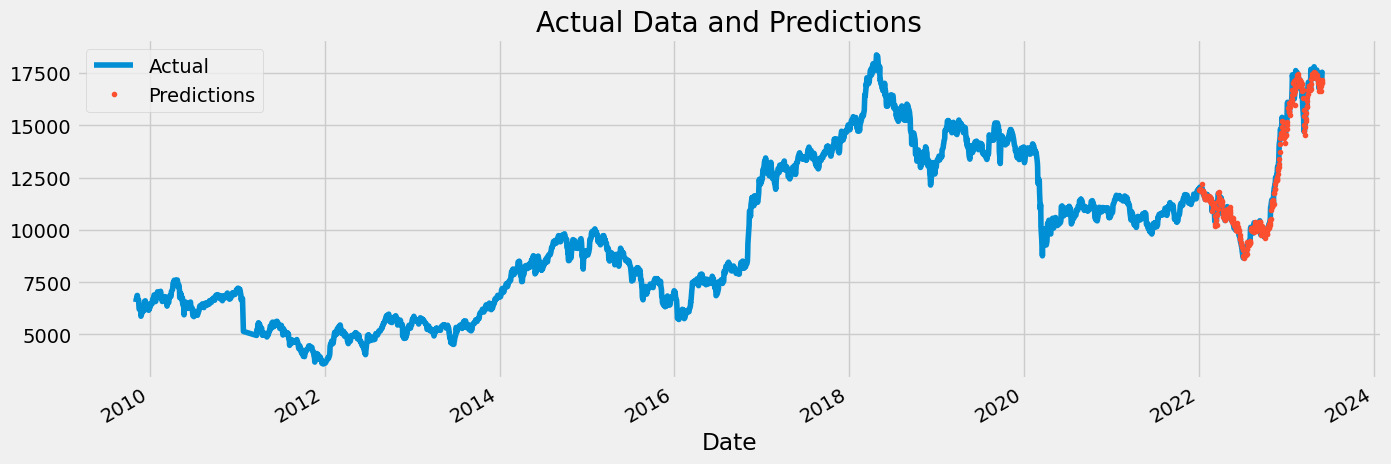}
    \caption{Comparison of predicted and real data over the time interval of one day according to the index price in points}
    \label{XGBOOST1}
\end{figure}

Robust performance across various metrics was observed in the assessment of single-day prediction models, with particularly noteworthy outcomes achieved by the XGBoost, LightGBM, and Random Forest models (see Table \ref{1day}). The most favorable results were demonstrated by XGBoost, indicating its superior predictive capabilities for the specific task. Comparable performance was noted for LightGBM, albeit with a marginally lower Mean Absolute Percentage Error (MAPE) compared to XGBoost. The efficacy of XGBoost and the commendable performance of LightGBM were emphasized, with a slightly diminished MAPE observed for the latter. The Random Forest model, known for its resilience and adept handling of intricate data relationships, showcased commendable predictive prowess; however, it was acknowledged that XGBoost and LightGBM surpassed it in performance metrics \ref{XGBOOST1}.
\begin{table}[!t]
\caption{Performance Metrics for Different Models (one-day)}
\label{1day}
\centering
\begin{tabular}{lccc}
\hline
\bfseries Model & \bfseries RMSE & \bfseries MAPE & \bfseries R2 \\
\hline\hline
XGBoost          & 0.0163 & 0.0187 & 0.993 \\
LightGBM         & 0.0163 & 0.019  & 0.993 \\
AdaBoost         & 0.0222 & 0.301  & 0.987 \\
KNN              & 0.0585 & 0.0799 & 0.908 \\
Decision Trees   & 0.0231 & 0.0266 & 0.986 \\
Random Forest    & 0.0165 & 0.0194 & 0.993 \\
Extra Trees      & 0.0172 & 0.205  & 0.992 \\
LSTM             & 0.0273 & 0.0325 & 0.982 \\
GRU              & 0.0228 & 0.0274 & 0.987 \\
\hline
\end{tabular}
\end{table}

\subsubsection{One-Week Predictions}

\begin{figure}
    \centering
    \includegraphics[width=1\linewidth]{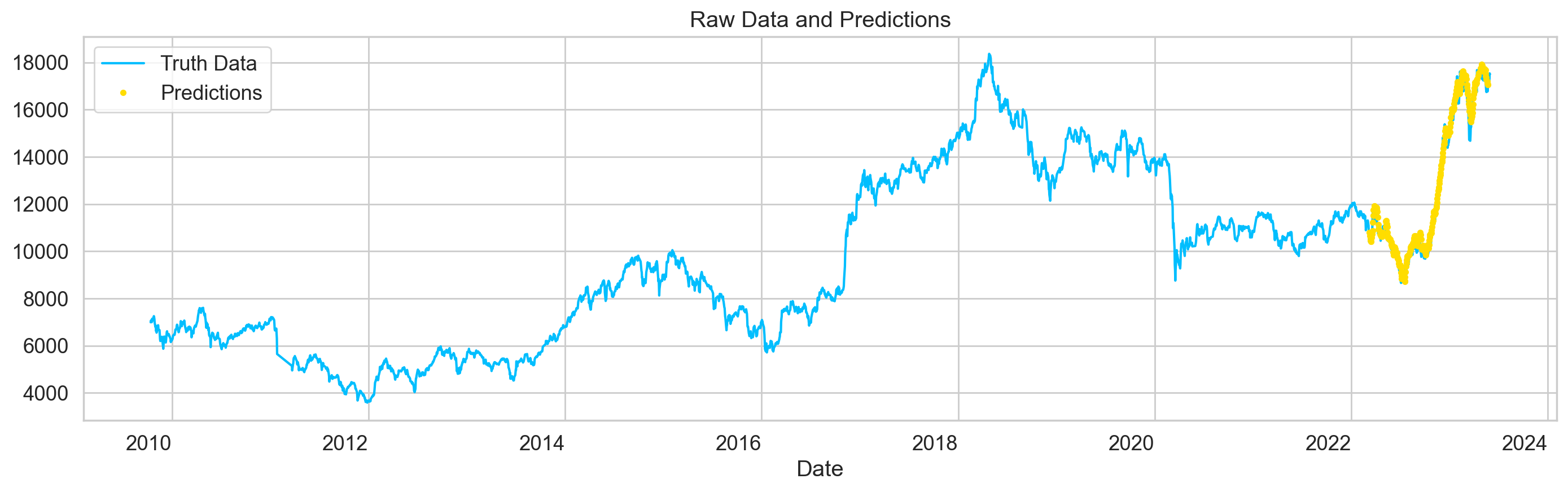}
    \caption{Comparison of predicted and real data over the time interval of one week according to the index price in points}
    \label{GRU5}
\end{figure}

Optimal performance in the assessment of one-week predictions was demonstrated by the Random Forest, Extra Trees, and GRU models(see Table \ref{1week}). The exceptional performance of the GRU model, marked by the attainment of the most favorable Root Mean Square Error (RMSE) of 0.0406 and a commendable R2 value of 0.961, was particularly noteworthy.
A noteworthy increase in Root Mean Square Error (RMSE) and Mean Absolute Percentage Error (MAPE) values, when compared to one-day predictions, was observed in extending the prediction horizon to one week. This increase underscores the heightened challenge and intricacy associated with forecasting over a more extended timeframe. The observed error elevation elucidates the intensified difficulty in accurately predicting over a more extended period.
The superior performance of the GRU model can be attributed to its adeptness in capturing seasonality and intricate relationships within the data. This capability positions GRU as a more effective model for addressing the augmented complexity inherent in one-week predictions, surpassing other models (See Fig. \ref{GRU5}).

\begin{table}[!t]
\caption{Performance Metrics for Different Models (one-Week)}
\label{1week}
\centering
\begin{tabular}{lccc}
\hline
\bfseries Model & \bfseries RMSE & \bfseries MAPE & \bfseries R2 \\
\hline\hline
XGB             & 0.0452 & 0.0589 & 0.946 \\
LGBM            & 0.0461 & 0.0641 & 0.944 \\
AdaBoost        & 0.0445 & 0.0563 & 0.947 \\
KNN             & 0.0691 & 0.0892 & 0.873 \\
DTR             & 0.0563 & 0.0784 & 0.916 \\
RF              & 0.041  & 0.0541 & 0.955 \\
Extra Trees     & 0.0412 & 0.051  & 0.955 \\
LSTM            & 0.047  & 0.0584 & 0.947 \\
GRU             & 0.0406 & 0.0535 & 0.961 \\
\hline
\end{tabular}
\end{table}

\subsection{Long Term Predictions}
\subsubsection{One-Month Prediction}
\begin{figure}
    \centering
    \includegraphics[width=1\linewidth]{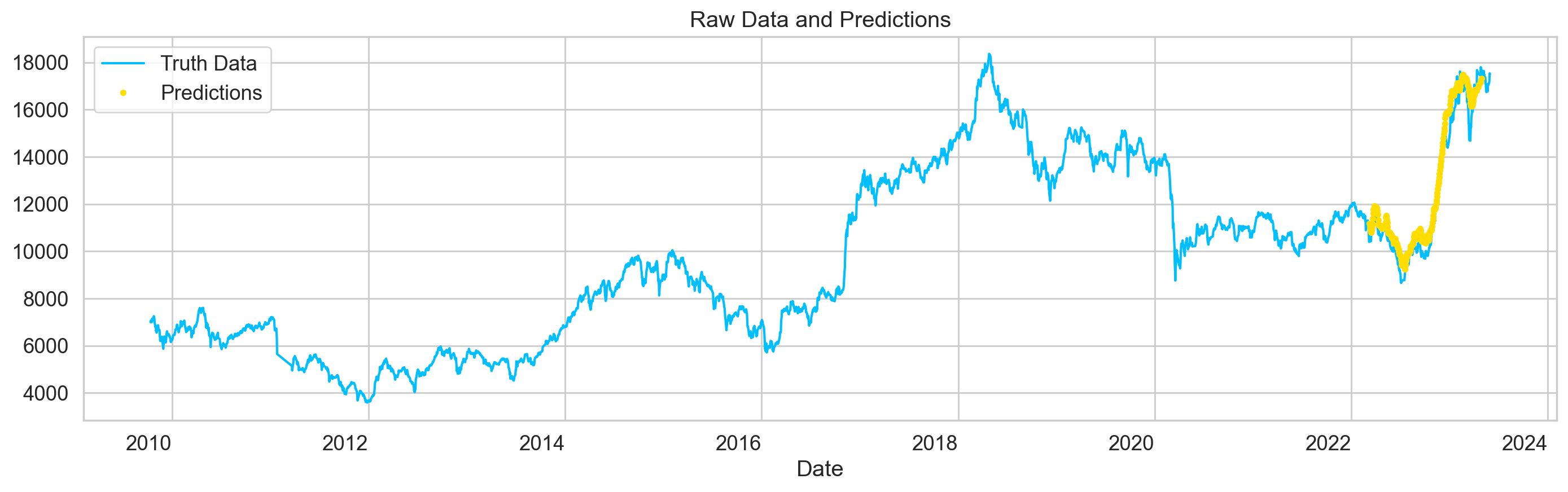}
    \caption{Comparison of predicted and real data over the time interval of one week according to the index price in points}
    \label{GRU22}
\end{figure}

In the context of our one-month prediction period, the performance analysis revealed intriguing outcomes for the LSTM, GRU, and unexpectedly, the KNN models (See Table \ref{1month}. Notably, the GRU model demonstrated superior performance across our three evaluation metrics, showcasing its efficacy in capturing intricate relationships and seasonality. The deep learning models, LSTM and GRU, particularly excelled in this extended forecasting horizon, underscoring their proficiency in handling complex temporal dependencies.
However, the most surprising revelation was the significant improvement in the performance of the KNN model. Despite its underperformance in the one-week and one-day tests, the KNN model emerged as the third-best performer in the one-month experiment. This unexpected success implies that KNN's nearest-neighbor approach may be well-suited for discerning specific characteristics inherent in monthly data. This emphasizes the critical importance of aligning model choices with the unique temporal and structural aspects of the forecasting task (See Fig. \ref{GRU22}.
\begin{table}[!t]
\caption{Performance Metrics for Different Models (1 Month)}
\label{1month}
\centering
\begin{tabular}{lccc}
\hline
\bfseries Model & \bfseries RMSE & \bfseries MAPE & \bfseries R2 \\
\hline\hline
XGB             & 0.201  & 0.224 & -0.0217 \\
LGBM            & 0.242  & 0.259 & -0.484 \\
AdaBoost        & 0.135  & 0.16  & 0.535  \\
KNN             & 0.107  & 0.13  & 0.712  \\
DTR             & 0.237  & 0.265 & -0.431 \\
RF              & 0.241  & 0.264 & -0.472 \\
Extra Trees     & 0.111  & 0.131 & 0.685  \\
LSTM            & 0.0965 & 0.118 & 0.788  \\
GRU             & 0.078  & 0.11  & 0.865  \\
\hline
\end{tabular}
\end{table}
\subsubsection{Two-Months Predictions}
\begin{figure}
    \centering
    \includegraphics[width=1\linewidth]{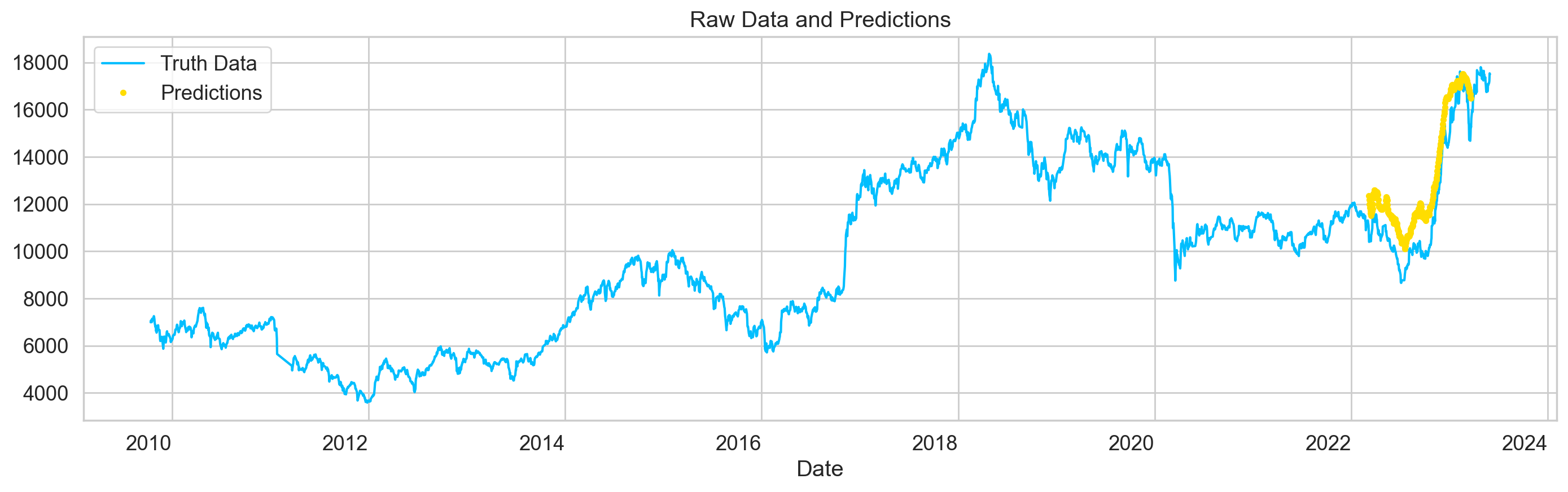}
    \caption{Comparison of predicted and real data over the time interval of one week according to the index price in points}
    \label{GRU44}
\end{figure}

When the analytical scope was extended to a two-month prediction period, consistency in top-performing models was observed, with the superior positions being maintained by the GRU, LSTM, and surprisingly, the KNN (see Table \ref{2months}. Excellence was sustained by the GRU, which exhibited the best Root Mean Square Error (RMSE) and R2 scores. The effectiveness in capturing temporal patterns by the deep learning models, namely GRU and LSTM, continued to be demonstrated, reaffirming their suitability for prolonged prediction horizons.
The unexpected yet consistent inclusion of the KNN model in both the one-month and two-month periods highlighted its adaptability to longer-term trends. This surprising presence underscored the capacity of the model to discern and leverage specific characteristics pertinent to extended forecasting durations.
The notable constancy in the performance of these models across varied prediction durations was emphasized, emphasizing their reliability and versatility for diverse forecasting timelines. The sustained effectiveness of GRU, LSTM, and KNN further underscored their resilience and applicability in addressing forecasting challenges over extended periods (See Fig. \ref{GRU44}).
\begin{table}[!t]
\caption{Performance Metrics for Different Models (2 Months)}
\label{2months}
\centering
\begin{tabular}{lccc}
\hline
\bfseries Model & \bfseries RMSE & \bfseries MAPE & \bfseries R2 \\
\hline\hline
XGB             & 0.255 & 0.267 & -0.558 \\
LGBM            & 0.27  & 0.29  & -0.754 \\
AdaBoost        & 0.177 & 0.191 & 0.246  \\
KNN             & 0.138 & 0.146 & 0.544  \\
DTR             & 0.269 & 0.276 & -0.738 \\
RF              & 0.273 & 0.291 & -0.781 \\
Extra Trees     & 0.168 & 0.175 & 0.32   \\
LSTM            & 0.134 & 0.142 & 0.605  \\
GRU             & 0.12  & 0.176 & 0.685  \\
\hline
\end{tabular}
\end{table}
\subsection{Ensemble Technique}
In addition to individual model evaluations, an ensemble technique was employed to further enhance predictive performance with main focus on the long term predictions (See Table \ref{Ensemble}).
The results revealed notably improved predictions for the long term compared to individual model evaluations. However, it's important to note that for single-day predictions, the accuracy of the ensemble was lower than that of XGBoost.This difference can be attributed to the  high fluctuations observed from day to day
The ensemble model exhibited the lowest Root Mean Square Error (RMSE) for the one-week, one-month and two-month prediction indicating its heightened capability in capturing long-term and short term trends with much better performance on the long term compared to individual models. In contrast, the GRU model, while performing better than the other models it gave over five times higher RMSE than the ensemble in the two-month prediction scenario. Its also noteworthy that as shown in Fig. \ref{Ensemble} the trend forcasting for the one month was not very good but for the rest (especially the two-months) it appeared to be very accurate with high ability to predict the ups and downs of the curve.
\begin{table}[!t]
\caption{Ensemble Blending Performance Metrics}
\label{Ensemble}
\centering
\begin{tabular}{lccc}
\hline
\bfseries Time Frame & \bfseries RMSE & \bfseries MAPE & \bfseries R2 \\
\hline\hline
1 Day   & 0.024  & 0.0215 & 0.934 \\
1 Week  & 0.0215 & 0.0205 & 0.953 \\
1 Month & 0.0284 & 0.0261 & 0.9   \\
2 Months& 0.0205 & 0.0175 & 0.651 \\
\hline
\end{tabular}
\end{table}
\begin{figure}
    \centering
    \includegraphics[width=1\linewidth]{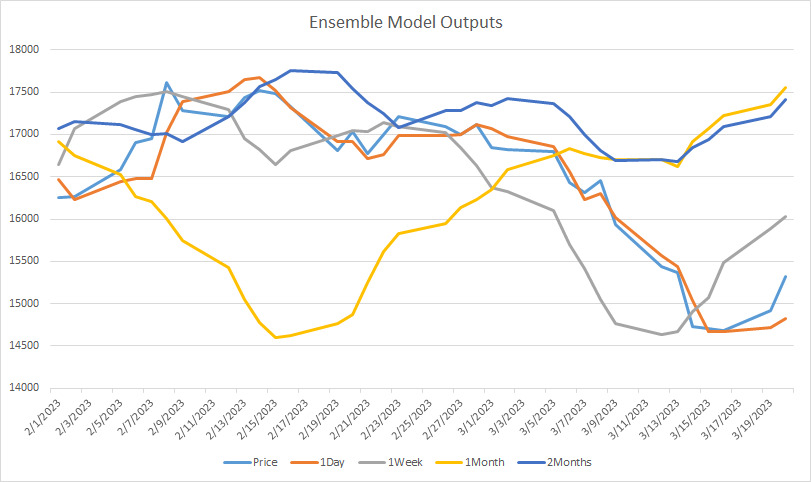}
    \caption{Comparison of predicted and real data over the time interval of one-day, one-week, one-month, and two-months according to the index price in points}
    \label{Ensemble}
\end{figure}
\subsection{Key Findings}
In the examination of forecasting models for EGX30 stock prices, the focus of this study was on evaluating the performance of diverse machine learning algorithms across varying time horizons. Notably, for single-day predictions, it was observed that XGBoost emerged as the most effective model, displaying superior accuracy in terms of Root Mean Square Error (RMSE), Mean Absolute Percentage Error (MAPE), and R-squared (R2) metrics.
As the forecasting horizon extended to one week, heightened efficacy was demonstrated by Random Forest, Extra Trees, and Gated Recurrent Unit (GRU). Exceptional proficiency in capturing complex relations and seasonality was showcased by the GRU model, consistently outperforming other models in extended periods.
Furthermore, an unforeseen strength in the k-Nearest Neighbors (KNN) model was revealed by our analysis. Traditionally under-performing in short-term predictions, the KNN model proved to be remarkably effective in capturing long-term trends. This unexpected outcome suggests the adaptability of the KNN model to specific characteristics and dependencies within monthly data.
\section{Conclusion} \label{Conclusion}
In conclusion, the imperative need to explore and enhance predictive models for stock prices in the Egyptian market, with a focus on the EGX30 index, is addressed in this study. A comprehensive comparative analysis of various machine learning models is conducted, including K-Nearest Neighbours, which surprisingly performed well with the long-term scope, Random Forest, Extreme Gradient Boosting showing excellent results in 1-day predictions, and Gated Recurrent Unit networks.
The methodology involves the use of lag functions for forecasting to identify the most efficient model in both the long and short terms. Evaluation metrics, such as root mean squared error, are utilized for model comparison. The findings underscore the effectiveness of machine learning techniques in predicting stock prices for the EGX30, highlighting the unique dynamics and traits specific to Egyptian markets.
It is observed that Gated Recurrent Unit Network (GRU) models demonstrate heightened efficacy compared to alternative methodologies in 1-week, 1-month, and 2-month predictions, while XGBoost outperforms it in 1-day predictions. Furthermore, the incorporation of an ensemble approach enhances predictive performance, showcasing its utility as a refined and synergistic predictive paradigm.
\section{Limitations \& Future Work} \label{Limitations and Futre Work}
Fine-tuning numerous models requires significant computational resources and time. Executing an effective fine-tuning loop is challenging which directly affecting the accuracy of the models. As part of future work, there is a plan to expand this analysis by incorporating additional machine learning (ML) and deep learning (DL) models. Specifically, the focus will be on integrating transforms, which have demonstrated notable performance in time series predictions. Also it is aimed to use the sentiment analysis to the business global and local calendars, news and public figures posts in social media.


\bibliographystyle{IEEEtran}
\bibliography{refrences}

\end{document}